# A Parallel/Distributed Algorithmic Framework for Mining All Quantitative Association Rules

Ioannis T. Christou, Athens Information Technology, Monumental Plaza, Bld. C, 44 Kifisias Ave., Marousi 15125, Greece, (ichr@ait.edu.gr, +30-210-668-2725) (Corresponding Author)

Emmanouil Amolochitis, Cepal Hellas Financial Services, Athens, Greece

Zheng-Hua Tan, Dept of Electronic Systems, Aalborg University, Aalborg, Denmark

**Abstract**—We present QARMA, an efficient novel parallel algorithm for mining all Quantitative Association Rules in large multi–dimensional datasets where items are required to have at least a single common attribute to be specified in the rules' single consequent item. Given a minimum support level and a set of threshold criteria of interestingness measures (such as confidence, conviction etc.), our algorithm guarantees the generation of all non-dominated Quantitative Association Rules that meet the minimum support and interestingness requirements. Such rules can be of great importance to marketing departments seeking to optimize targeted campaigns, or general market segmentation; they can also be of value in medical applications, financial as well as predictive maintainance domains. We provide computational results showing the scalability of our algorithm, and its capability to produce all rules to be found in large–scale synthetic and real–world datasets such as Movie–Lens, within a few seconds or minutes of computational time on commodity off–the–shelf hardware.

## 1 INTRODUCTION

Quantitative association rules (QAR) [1,2,3] form an integral part of Association Rule Mining [3,4,5]; nevertheless, most of the literature on algorithms for the discovery of association rules in databases focuses on Boolean association rule mining, that is, finding all rules with certain support and confidence levels of the form $A \to B$ where $A, B$ are subsets of the entire set of items (inventory catalogue) $S$ that appear in a database $D$ of historical transactions. Such rules are qualitative (Boolean) in nature as no attribute–related information of the items in the antecedent or the head of the rule are taken into account, and all that matters is the existence or not of the items in question in a transaction, even though certain information, such as the price paid for each item, could be of great importance.

In our context for Quantitative Association Rule Mining (QARM), the database $D$ comprises of historical transactions of users purchasing items from a collection $S$ of items, each characterized by a set of categorical ($A_C^i$) and quantitative attributes ($A_Q^i$), together forming the set of attributes $A^i = A_C^i \cup A_Q^i$ for the item $i$; and for each attribute $a \in A^i$, depending on whether it's a categorical or quantitative attribute, there exists an associated set of possible values $R^a$ the attribute may take, called the attribute's *range*, so that

$$R^a = \begin{cases} [l^a, u^a] \subseteq \overline{\mathbb{R}} = [-\infty, +\infty], & a \in A_Q^i \\ \{v_1^a, \dots, v_n^a\}, & a \in A_C^i \end{cases}$$

The database $D$ is arranged in records of user histories $H$ containing information about every item purchased by each user. A user history therefore is the set of all transactions made by that particular user. A single transaction is a record containing a unique identification number of the user who made the transaction, and also information about the single item in that transaction: the item's id, together with values for one or more of the item's attributes; for example, the item's purchase price. We denote by $S^i$ the set of all user histories containing at least one instance of item $i$. The QARM task is then to discover *all interesting* quantitative rules of the form

$$i_1[v^{a_{i_{1,1}}} \in r_{i_{1,1}}] \wedge \dots \wedge i_n[v^{a_{i_{n,k}}} \in r_{i_{n,k}}] \underset{s,c}{\to} j_1[v^{a_{j_{1,1}}} \in r_{j_{1,1}}] \wedge \dots \wedge j_m[v^{a_{j_{m,l}}} \in r_{j_{m,l}}] \qquad (1)$$

where the notation $i_j[v^a \in r]$ (where $a \in A^{i_j}$) denotes the "restriction" of the records in $H$ containing item $i_j$ to those ones for which each of the attributes $a \in A^{i_j}$ takes values $v^a$ from the set $r \subseteq R^a$. The notation in equation (1) should be interpreted as "IF the value of attribute $a_{i_{1,1}}$ for item $i_1$ is in the set $r_{i_{1,1}}$, ... and the value of attribute $a_{i_{n,k}}$ for item $i_n$ is in the set $r_{i_{n,k}}$, THEN, with *interestingness* level(s) $c$, and support $s$, item $j_1$ appears in the same record and the value of its attribute $a_{j_{1,1}}$ is in the set $r_{j_{1,1}}$, and ... and item $j_m$ appears in the same record and the value of its attribute $a_{j_{m,l}}$ is in the set $r_{j_{m,l}}$".

Such rules are in fact *multi–dimensional quantitative association rules* [6 pp. 251—257], as there are many distinct attributes (dimensions) involved in the definition of the same rule. By "interestingness" of a rule, we mean that in the dataset, a rule must take on values for a set of defined metrics above or below specified thresholds. In usual practice, the single metric defining interestingness of a rule is the *confidence* (that must be higher than specified threshold); however, other metrics or combination thereof might equally well be applied within our system: for example, in the work we present it is possible to define as interestingness criteria of a rule that a rule must have (a) confidence above a minimum confidence threshold and (b) absolute value of leverage [3] above a minimum leverage threshold. We will define the notions of support, confidence and conviction [7], [8 pp. 151—159] for quantitative rules as a natural (minor) extension of the corresponding notions of support, confidence & conviction for the standard qualitative association rules, in the subsequent sections. The problem has been shown to be in general NP–hard [4, 9]. In this work, we present QARMA, a fast parallel exact algorithm that computes all non-dominated (notion precisely defined in section 2.1, see Definition 3) quantitative association rules with minimum support and interestingness of the slightly restricted form

$$i_1[p \geq l_{p1}] \wedge i_1[v^{a_{i_1}} \geq l_{a_{i_1}}] \wedge \ldots \wedge i_n[p \geq l_{pn}] \wedge \ldots \wedge i_n[v^{a_{i_{n,m}}} \geq l_{a_{i_{n,m}}}] \underset{s,c}{\rightarrow} j[p \gtreqqless l_j] \qquad (2)$$

i.e. we require that there is exactly one item specified in the rule's consequents, that there exist at least one shared quantitative attribute $p$ for each item in the dataset, and we consider the problem of mining (multi–dimensional) quantitative association rules whose items in the consequent constrain only the value of this shared attribute to intervals of the form $[l, +\infty)$ or alternatively, to exact values $\{l\}$ (the symbol $'\gtreqqless'$ in eq. (2) is to be interpreted as "exclusively either '$\geq$' or else '='".) In the antecedent, there may be any combination of attributes in any of the items, taking values on left-closed, half-open intervals of the previous form. This restriction does *not* prevent the algorithm in any way from discovering rules that constrain the attribute values of the antecedent items in *fully closed intervals*, as the introduction of an artificial attribute that takes on values equal to the negative of the original attribute value results in the discovery of all rules where the intervals on the antecedent items' attributes may take on any form (half–open on any end, or fully closed intervals, or even singleton value intervals). Indeed, it is trivial to see that the constraint $i[v^{a_i} \in [l, u]]$ is equivalent to the expression $i[v^{a_i} \geq l] \wedge i[\langle -v^{a_i}\rangle \geq -u]$ where $\langle -v^{a_i}\rangle$ denotes the artificial attribute just discussed. By having $l = u$ we see we also cover the singleton value interval case. Therefore, the rules our algorithm discovers are constrained from the general multi–dimensional form (1) only in that a shared attribute must be present in all items in the dataset and that the right–hand–side of the rules may only specify half-open (closed–left) intervals (or single values) for the single shared attribute among all items. In the following, we will consider the setting where the quantitative attribute in the rule's consequent item must take values on left–closed half–open intervals only, and we will discuss the case of exact single values in section 3.4, together with some other possible extensions/modifications; in that section we also show through computational results the scalability of QARMA. We discuss the applications of this algorithm in a number of domains, and provide quantitative results showing the feasibility of our approach even for large–scale datasets.

## 1.1 Motivation

A major motivating factor for this research has been our work on movie recommendation systems for triple–play services providers [10–12]. In many such settings (see also [13–14]), the price for a movie is a function of time (items are depreciated over time), and plays a very important role in the final decision of a user regarding the purchase or not of an item; indeed, market research [15] shows that currently, more than 90% of all items viewed by video–on–demand (VoD) subscribers belong to the "free item" category, and that the average subscriber in the EU pays just under two Euros per month on VoD rentals over and above their fixed monthly subscription fees. Yet, recommender systems algorithms do not systematically exploit this fact in order to either increase their recall rates and subsequently increase customer retention rates, or, even better, understand for each of their customers individually, their "ideal" price-range for a movie, and then potentially make a more appealing offer to each customer. In the context of recommender systems, we use our algorithm in two ways: (a) to build a post–processing tool that re–ranks the top-n recommendations of another recommender engine [10–11] based on the rules found by the algorithm; and (b) to build a personalized reservation price estimator that computes an approximation of a given customer's reservation price regarding a given item, so as to possibly adjust the price of that item specifically for that customer alone. The second way in which we use our algorithm (so as to obtain first–degree pricing differentiation, see [16]), can have a significant impact for e-commerce, and commerce in general, as our computational results show.

## 1.2 Related Work

Piatetsky–Shapiro [3] presented the first algorithm for Quantitative Association Rule Mining in 1991, even though the exposition was for single antecedent and single consequent attributes. Srikant & Agrawal [1] showed how to overcome the problem of finding summaries for all combinations of attributes (which is exponentially large) required for a straight–forward extension of Piatetsky–Shapiro's algorithm to multiple antecedent and/or consequent attributes in the rules by a decomposition/partitioning of the quantitative attributes' intervals followed by possible merging and pruning the search space of candidate itemsets. Salleb–Aouissi et al. [17] developed QuantMiner, a GA for mining interesting QARs via an optimization process; the idea of optimizing QARs originally appeared in Fukuda et al. [18] where however the algorithms presented aim to find intervals for quantitative attributes that maximize support and/or confidence of the produced rules. More recent work on finding QARs using evolutionary algorithms is presented in [19] where the authors used multi-objective Genetic Algorithms incorporating in their fitness function the notion of rule interestingness, and similarly in [20], where Alvarez and Vasquez developed an evolutionary algorithm to discover interesting QARs without any a–priori discretization. The work of Ruckert et al. [21] is also related to our work; in this paper the authors proposed looking for half-space conditions on the quantitative attributes instead of hyper-rectangular representations. In their work, the authors seek to produce rules where the antecedent or the consequent are formalized as linear inequalities of the quantitative attributes, which includes the rules we search for as special cases. However, instead of seeking to produce all such interesting rules that exist, the authors resort to a gradient-descent-based algorithm for producing so-called locally optimal rules (optimal according to their notion of "rule interestingness"), and they give no guarantees as to the quality of the solutions their algorithm finds.

Finally, regarding the application of standard association rule mining in the field of recommender systems, a first approach to collaborative filtering-based recommendation via adaptive association rule mining is presented in [22]. Ye [23] presented a system that combines association rule mining and self–organizing maps to create personalized recommendations. Leung et al. [24] also developed a collaborative filtering framework using fuzzy association rules incorporating multi–level similarity measures that improved prediction quality.

### 1.3 Our Contribution

We present QARMA (short for QAR Mining Algorithm), a new highly parallelizable algorithm for mining all interesting quantitative association rules, that is, non–dominated rules having support above a minimum support threshold $s_{min}$, and for a set of interestingness metrics (e.g. confidence) have levels above minimum specified threshold values $c_{min}$. The algorithm works on datasets where all items share at least one attribute (e.g. price), and produces rules that specify values for the common attribute of the consequent item in a left–closed interval. The rules also specify minimum values for each specific attribute of each item in the antecedents of the rule. Our algorithm is significantly different from other algorithms as it does not use any value discretization or bucketing technique (though such techniques may easily be incorporated and even be needed when dealing with dense datasets whose attributes take very many different numeric values: see section 3.4), and in addition guarantees the generation of all such rules. We prove its importance in real–world business decision making problems.

In section 3.2 we run experiments with a synthetic e-commerce dataset. In section 3.3 we run experiments on another synthetic dataset to test whether QARMA can detect rare event conditions in predictive maintenance settings. In section 3.4 we run experiments with the Movie–Lens dataset ml-1m. We show the scalability of QARMA in terms of the speed–up obtained by using more processing units, and we also make comparisons against other QARM algorithms.

## 2 THE QARMA ALGORITHM

### 2.1 Definitions

In the following, we assume that $Q$ denotes a finite set of tuples of the form $(i, a, v)$ where $i \in S$, $a \in A_Q^i, v \in \mathbb{R}$. The pair $(i, a)$ forms a key of that set so that no two tuples with the same item and attribute can exist in $Q$. We denote a quantitative association rule of the form (2) as $r = (B \to I|Q)$ where $B \subset S, I \in S$, with the following interpretation: with sufficiently high support and interestingness, the existence of all items in $B$ in a user history $t$ such that for each item $i \in B$ that appears in any pair $(i, a, v) \in Q$ there exists a transaction in $t$ for that item with value for the attribute $a, v_a \geq v$ imply the existence of $I$ in that user's set of historical transactions, and that the value of the common shared attribute $p$ of item $I$ is *greater than or equal* to the value specified in the pair $(I, p, v) \in Q$. Due to our imposed restriction of mining rules specifying values for only a particular user-defined attribute $p$ in the consequent item of the rule, we have the following property of the tuples in $Q$: $\forall r = (B \to I|Q): ((I, a, v) \in Q \to a = p)$. For a given dataset $D$ we denote the set of all possible quantitative rules in that set by $R^D$ (not to be confused with the $d$-dimensional real space $\mathbb{R}^d$.)

**Definition 1. (Rule Confidence)** A rule $r = (B \to I|Q)$ has **confidence** $c$ denoted by $CONFIDENCE(r)$ that equals the number of user histories that contain each of the items in $B \cup \{I\}$ with attribute values for each of the specified attributes in $Q$, at least equal to the value specified for that item–attribute pair in $Q$ **divided** by the number of user histories in which each of the items in $B$ appears with attribute values for each of the specified attributes in $Q$, at least equal to the value specified for that item–attribute pair in $Q$. If an item does not appear in any of the pairs in $Q$, the condition is reduced to the mere existence of the item in one or more transactions of the user history, regardless of any attribute value.

**Definition 2. (Rule Support)** A rule $r = (B \to I|Q)$ has **support** $s$ denoted by $SUPPORT(r)$ that equals the number of user histories in which each of the items in $B \cup \{I\}$ appears with attribute values for each of the specified attributes in $Q$, at least equal to the value specified for that item–attribute pair in $Q$ **divided** by the total number of user histories in the database. If an item does not appear in any of the pairs in $Q$, the condition is reduced to the mere existence of the item in one transaction of the user history, regardless of any attribute value.

**Definition 3. (Rule Conviction)** A rule $r = (B \to I|Q)$ has **conviction** $t$ denoted by $CONVICTION(r)$ that equals the ratio $(1 - CONS\_SUPP(r))/(1 - CONFIDENCE(r))$ where the consequent support $CONS\_SUPP(r)$ is the number of user histories in which the item $I$ appears with consequent attribute value at least equal to the value specified for that item–attribute pair in $Q$ **divided** by the total number of user histories in the database (if the consequent item does not appear in any of the pairs in $Q$, the condition is reduced to the mere existence of the item in one transaction of the user history, regardless of attribute value.)

**Definition 4. (Rule Dominance)** Given a transitive predicate formula $LTF(r, r')$ derived from a finite set of interestingness metrics $M = \{F: R^D \to \mathbb{R}\}$ satisfying $\forall r, r', r'' \in R^D: LTF(r, r') \land LTF(r', r'') \to LTF(r, r'')$, we say that a rule $r = (B \to I|Q)$ is dominated by another rule $r' = (B' \to I|Q')$, and we write $r' > r$ if:

$B \supseteq B'$ AND

$(I, p, v) \in Q, (I, p, v') \in Q' \to v' \geq v$ (i.e. the consequent's attribute value in $r'$ is greater than or equal to the consequent's attribute value in $r$) AND

$SUPPORT(r) \leq SUPPORT(r')$ AND

$LTF(r, r')$ AND

$\forall (K \in B', a \in A_Q^K, w^a) \in Q' \to \exists (K \in B, a \in A_Q^K, v^a) \in Q, w^a \leq v^a$.

The predicate that we shall assume by default will be $LTF(r, r') = CONFIDENCE(r) \leq CONFIDENCE(r')$; in another setting we could have $LTF(r, r') = CONFIDENCE(r) \leq CONFIDENCE(r') \land CONVICTION(r) \leq CONVICTION(r')$.

The above dominance definition holds when the rules' right-hand-side are inequalities; when mining rules where the rules' right-hand-side represent equalities, then the rule dominance criterion must be slightly modified to indicate that the dominant and dominated rule specify exactly the same value for their consequent item common attribute. There is no dominance relationship between two rules whose right–hand–sides are equalities, but specify different values each for their consequent item attribute.

**Definition 5. (Rule Coverage)** The covering set of a quantitative rule $r = (B \to I|Q)$ is defined as the subset of user-histories in the dataset for which the rule holds: $COV(r) = \{t \in D: (\forall (i \in B, a \in A_Q^i, v) \in Q): (\exists (i, a, v') \in t: v' \geq v) \land ((I, p, v) \in Q \to \exists (I, p, v') \in t: v' \geq v)\}$. The covering set of a set of rules $RS \subseteq R^D$ is defined as the union of the covering sets of each member of that rule–set: $COV(RS) = \cup_{r \in RS} COV(r)$. From the definition of support, we have $SUPPORT(r) = |COV(r)|/|D|$.

Notice that while our definitions of support and interestingness of a rule are direct applications of the respective notions in the realm of binary association rule mining, we also have to introduce the notion of dominance defined in Definition 4 above, so as to *discard valid* (satisfying minimum support and interestingness) *but useless rules*: a dominated rule is of no use as there exists another rule with equal or higher support and interestingness metrics that implies the same item as the dominated rule with equal or higher attribute value, and in addition, the dominated rule specifies higher values for each of the antecedent items' values specified in the other (dominant) rule. As a result, the dominated rule provides no new information, in that, whenever the dominated rule fires and signals a value for the common quantitative attribute of the consequent item $I$ higher than a value $v$, the dominant rule also fires, and signals an at least as high value of $I$'s attribute (providing equal or more information), with equal or higher support and interestingness values. A nice property of the notion of rule dominance is that of *transitivity*: $r_1 \succ r_2 \land r_2 \succ r_3 \Rightarrow r_1 \succ r_3$.

Another relation closely associated to the dominance relation between rules, is the *widening* relation between rules, that we define as follows: we say that a rule $r'$ is *wider* than a rule $r$ and we write $r' \sqsupset r$ if all the conditions of Definition 4 hold with the possible exception of the conditions for the relationship between the support or interestingness values of the two rules. It is easy to see that the widening relation between rules is also transitive, and that clearly the implication $r' \succ r \Rightarrow r' \sqsupset r$ holds, though the reverse implication is not true in general. It is easy to modify the QARMA algorithm to produce all the widest, non-dominated rules that exist for a given dataset, sometimes significantly reducing the total number of rules produced (the rationale being that as long as the rules have minimum support and interestingness, we don't care for rules for which there exist others wider than them in the result.)

Finally, there is the issue of *rule generalization*: given a valid non–dominated rule $r = (B \to I|Q)$ that meets minimum support and interestingness criteria on a given training dataset $D$, the evidence on that dataset also supports the following generalizations on the restrictions of the rule's item–attribute quantifications:

$\forall i \in B \cup \{I\}: (i, t, l_{i,t}) \in Q: i[v^{a_{i,t}} > p_{a_{i,t}}^-]$ where $p_{a_{i,t}}^-$ is the greatest value that attribute $t$ of item $i$ takes on the dataset $D$ that is strictly less than $l_{i,t}$ unless there is no such value in the dataset, in which case the restriction remains interpreted as $i[v^{a_{i,t}} \geq l_{i,t}]$.

## 2.2 Algorithm Specification

Our algorithm for computing all QARs from a transactional database expects a set of *user histories* (of the form {UserID:$X$, Items–Purchased: SET–OF (ItemID: $Y$, Attr: $W$, AttrValue: $Z$)}). As a concrete example, in the *single–attribute* case, we interpret QARs of the form (2) as:

"IF a user has purchased products $A_1 \ldots A_k$ for which, the corresponding attribute values are $p_{A1} \ldots p_{Ak}$ or higher, THEN such a user would be willing to purchase a product $C$ whose attribute value is at least $p_C$."

The above definition of what constitutes the input database allows for *multiple* "purchases" of the same item by the same user, which is very common in most cases our algorithm is designed for application. As an example, in e–commerce datasets, the same customer over a period, purchases in multiple transactions many copies of the same product. Also, in patient medical records, the same patient over a period of time, has the same medical test (e.g. blood pressure measurement) multiple times, and so on. Nevertheless such a feature is seldom found in other implementations, which usually assume that the input dataset consists of a single large table, in which in each row, each column may take on a single value (which in relational databases design, corresponds to the 1st normal form). Items are assumed to be in a transitive order relationship, given by a map $ord: S \to \mathbb{N}$ (that can be the sequence id by which items are stored in the db.) Similarly, attributes within an item are assumed to be in a transitive order relationship given by a map $ord: A_Q \times S \to \mathbb{N}$.

**QARMA Algorithm**
**INPUTS**: Database $D$, minimum required support $s$, set of interestingness metrics $M \subset \{F: R^D \to \mathbb{R}\}$, minimum required interestingness thresholds $c(m) \forall m \in M$, transitive predicate formula $LTF(r, r')$ defining when a rule is more interesting than another, consequent items' shared common attribute $p$, ordering maps $ord: S \to \mathbb{N}$ and $ord: A_Q \times S \to \mathbb{N}$.
**OUTPUTS**: All non-dominated QARs of the above form with support, interestingness above the minimum specified thresholds.
**Begin**
1. run procedure INIT($D, s, M, c(.), LTF, p$) to produce set of all frequent item-sets $F_s$ and map $P$.
2. **let** $R = \emptyset$.
3. **foreach** $k = 2 \ldots max_{\ell \in F_s} \|\ell\|$ **do**:
   3.0. **foreach unit of execution do**:
     3.0.1. Atomically get global read-lock.
     3.0.2. **set** $R_{local}$ to be a copy of $R$ local to the current unit of execution.
     3.0.3. Atomically release global read-lock.
   3.1. **endfor.**
   3.2. **parallel foreach** frequent $k$-itemset $\ell_k \in F_s$ **do**:

   3.2.1. **set** $H_1 = \{r = (B \to I): B = \ell_k \setminus \{I\}, I \in \ell_k\}$.
   3.2.2. **foreach** rule $r = (B \to I) \in H_1$ **do**:

     3.2.2.1. **let** $T = \emptyset$. // $T$ is a FIFO queue of sets of tuples of the form $\left(K \in \ell_k, a \in A_Q^K, p \in \{p_{K,a,1}, \ldots p_{K,a,n_{K,a}}\}\right)$
     3.2.2.2. **foreach** $i = n_{I,p}\left(= dim(P(I, p))\right) \ldots 1$ **do**:

       3.2.2.2.1. **let** $Q = \{(I, p, p_{I,p,i})\}$. // set the value of the common shared attribute $p$ of $I$ to $p_{I,p,i} = P(I, p)_i$.
       3.2.2.2.2. **if** $SUPPORT(B \to I|Q) < s$ **continue**.
       3.2.2.2.3. Insert $Q$ onto $T$.
       3.2.2.2.4. **while** $T \neq \emptyset$ **do**:
         3.2.2.2.4.1. Remove the first $Q$ from $T$.

         3.2.2.2.4.2. **foreach** $J \in B$ **do**:

           3.2.2.2.4.2.0. **if** $ord(J) < max_{L \in B, a \in A_Q^L}\{ord(L): \exists v: (L, a, v) \in Q\}$ **continue**. // avoid duplicate set creation

           3.2.2.2.4.2.1. **foreach** $a \in A_Q^J$ **do**:
             3.2.2.2.4.2.1.1. **if** $ord(J) < max_{A \in B}\{ord(A): \exists v: (A, a, v) \in Q\}$ **OR**

                      $ord(a, J) \leq max_{t \in A_Q^J}\{ord(t, J): \exists v: (J, t, v) \in Q\}$ **continue**. // avoid duplicate set creation
              3.2.2.2.4.2.1.2. **foreach** $j = 1 \ldots n_{J,a}\left(= dim(P(J, a))\right)$ **do**:

                3.2.2.2.4.2.1.2.1. **let** $Q' = Q \cup \{(J, a, p_{J,a,j})\}$.
                3.2.2.2.4.2.1.2.2. **if** $SUPPORT(B \to I|Q') \geq s$ **then**
                  3.2.2.2.4.2.1.2.2.1. Insert $Q'$ onto $T$.
                  3.2.2.2.4.2.1.2.2.2. **if** $\forall m \in M: m(B \to I|Q') \geq c(m)$ **then**
                    3.2.2.2.4.2.1.2.2.2.1. **if** $\neg$(Exists-dominating-rule-for$((B \to I|Q'), R_{local}, LTF))$ **then**
                      3.2.2.2.4.2.1.2.2.2.1.1. Add $(B \to I|Q')$ to $R_{local}$.
                  3.2.2.2.4.2.1.2.2.2.2. **endif** // no dominating rule exists
                  3.2.2.2.4.2.1.2.2.3. **endif** // interestingness metrics ok
                3.2.2.2.4.2.1.2.3. **else break**. // support not ok
                3.2.2.2.4.2.1.2.4. **endif**. // support
              3.2.2.2.4.2.1.3. **endfor**. // $j$
            3.2.2.2.4.2.2. **endfor**. // $a$
          3.2.2.2.4.3. **endfor**. // $J$
       3.2.2.2.5. **endwhile**. // $T$
     3.2.2.3. **endfor**. // $i$
   3.2.3. **endfor.** // rule loop over $r$
 3.3. **endfor parallel**. // loop over $\ell_k$
 3.4. **foreach unit of execution do:**
   3.4.1. Atomically get global write-lock.
   3.4.2. **set** $R = R \cup R_{local}$.
   3.4.3. Atomically release global write-lock.
 3.5. **endfor.**
4. **endfor**. // $k$
5. **return** $R$.
**End**

The procedure INIT() that is invoked as the first step of the QARMA algorithm (step 1.) is as follows:

**Procedure INIT**
**INPUTS**: Database $D$, minimum required support $s$, set of interestingness metrics $M \subset \{F: R^D \to \mathbb{R}\}$, minimum required interestingness thresholds $c(m) \;\forall m \in M$, transitive predicate formula $LTF(r, r')$ defining when a rule is more interesting than another, consequent items' shared common attribute $p$.

**OUTPUTS**: all frequent itemsets $F_s$ with support $s$ or higher (assuming minimum attribute value level for all attributes, i.e. by treating the database as qualitative database, all attribute values were at minimum). Map $P: S \times A_Q \to \bigcup_{k=1}^{max(n_{i,a}: i \in S, a \in A_Q^i)} \mathbb{R}^k$ with the property that $P(i, a)$ is a lexicographically sorted vector in $\mathbb{R}^{n_{i,a}}$, and clearly, there are $n_{i,a}$ distinct values in the database that the attribute $a$ takes for item $i$.
**Begin**

0. Compute all different values of each attribute $a \in A_Q^i$ of every item $i$ with which it occurs in $D$, and sort them in increasing order $0 \leq p_{i,a,1} < p_{i,a,2} ... < p_{i,a,n_{i,a}}$ and store them in map $P: S \times A_Q \to \bigcup_{k=1}^{max(n_{i,a}: i \in S, a \in A_Q^i)} \mathbb{R}^k$ so that $P(i,a)$ is a lexicographically sorted vector in $\mathbb{R}^{n_{i,a}}$, and clearly, there are $n_{i,a}$ distinct values in the database that the attribute $a$ takes for item $i$.

1. Use any algorithm for frequent itemset generation (such as the FP-Growth algorithm) to generate all frequent itemsets $F_s$ with support $s$ or higher (assuming minimum attribute value level for all attributes, i.e. by treating the database as a qualitative database).

**End**

The operation "**Exists-dominating-rule-for**$((B \to I|Q'), R_{local}, LTF)$" returns true if and only if the QAR $(B \to I|Q')$ is dominated by at least one other rule in the current QAR–set $R_{local}$ according to the predicate $LTF$.

The tests in step 3.2.2.2.4.2.0 $(ord(J) < max_{L \in B, a \in A_Q^L}\{ord(L): \exists v: (L,a,v) \in Q\})$ and 3.2.2.2.4.2.1.1 $(ord(J) < max_{A \in B}\{ord(A): \exists v: (A,a,v) \in Q\}$ OR $ord(a,J) \leq max_{t \in A_Q^J}\{ord(t,J): \exists v: (J,t,v) \in Q\})$ need some further explanation.

The 1$^{st}$ check essentially verifies that given the partially quantified rule $r = (B \to I|Q)$ under consideration (after the "popping" of the oldest set $Q$ from $T$ in step 3.2.2.2.4.1), the item $J \in B$ to be considered for quantifications of its attributes is either the one with maximum id (according to the predefined item-order function $ord(.)$) quantified for any of its attributes, or else has not been quantified at all yet. This check avoids computing quantifications that have already been tested before or that will be tested in a subsequent step: all the possible quantifications of all the attributes of item $J$ are tested when this item becomes the one among antecedent items to enter the quantifications of the rule. To make this point more clear, consider an example un-quantified rule $a \wedge b \wedge c \to d$ and assume that items are ordered in lexicographic order. Given the set $Q = \{(a,p,1), (c,p,2), (d,p,1)\}$, the algorithm in step 3.2.2.2.4.0 will correctly decide not to consider any quantifications for the item $b$ because they will lead to quantifications that will be the same as the quantifications of the attributes of item $c$ when starting with partial quantifications $Q$ of attributes of the items $a, b$.

And the 2$^{nd}$ check also tests that while quantifying the attributes of an antecedent item, the algorithm does not attempt to set again an attribute which has been set in a previous iteration of the loop. The check makes sure that only attributes $a$ of item $J$ that have not yet been quantified in the current set $Q$ are eligible for quantification. Again, to make this point clearer, similar to the case for the 1$^{st}$ check, consider an item $a$ with three attributes $t_1, t_2, t_3$ besides the common attribute $p$ (ordered as $p, t_1, t_2, t_3$) and consider the partially quantified rule $a[p \geq 1][t_1 \geq 0][t_3 \geq 1] \to b[p \geq 5]$ represented by the set $Q = \{(a,p,1), (a,t_1,0), (a,t_3,1), (b,p,5)\}$. The algorithm in step 3.2.2.2.4.2.1.1 will correctly decide not to consider any quantifications for any attribute of item $a$. Quantifications of the yet unquantified attribute $t_2$ would have happened earlier when considering the set $Q' = \{(a,p,1), (a,t_1,0), (b,p,5)\}$, and if any quantification of the attribute also passed the support test for some value $X$, an augmented set $Q'' = \{(a,p,1), (a,t_1,0), (a,t_2,X), (b,p,5)\}$ would then have entered the queue $T$. And when testing this latter set, the algorithm would then examine the possibility of quantifying the last non-quantified attribute of item $a$ (attribute $t_3$).

Notice that the data structures created in steps 3.2 – 3.3 such as the queue $T$ or the QAR–set $R_{local}$ are all *local* to the units of execution (in their private memory), and are not accessible to other units of execution or even to a subsequent iteration of that loop, which enables us to use the Object-Pool pattern in a thread-local manner [25]. In order to speed up the operation "Exists-dominating-rule-for(.,.)", the data structure containing the set $R_{local}$ is implemented with the help of a hash-table that hashes the consequent item of a rule to linked lists of rules having the same consequent, in order of decreasing attribute value specified for the consequent. Insertion of a new rule into the QAR–set is then a matter of scanning the right list until the proper place to insert the rule is found, or a dominating rule is found before. As for the computation of the support and interestingness metrics, they can be made very fast if appropriate hash-tables are created in a pre-processing step that store for each item and attribute–level pair, a bit–vector containing the (consecutive) ids of all users that have in their histories the item specified with an attribute value level higher than the one specified in the key. With such a cache, obtaining for example the support of a rule is a simple matter of AND-ing together all bit–vectors of all item-attribute value quantifications in the rule, and counting the number of bits set in the resulting bit–vector.

Finally, notice that the algorithm as described is especially well–suited to a *shared–memory multi–processing model which is commonly found in today's multi/many–core computers*, and the read/write locks required are currently built–in standard library methods for most modern programming languages offering native support for multi–threading including Java and C++. If however one wishes to utilize a distributed memory multi-processor system such as a cluster of interconnected multi–core servers, then a distributed memory global read/write lock mechanism or a distributed barrier that allows synchronization across processes and address spaces has to be utilized; both such mechanisms based on passing "active" messages are found in the "`parallel.distributed`" package of the Open-Source popt4jlib Java library https://github.com/ioannischristou/popt4jlib developed by the first author.

As a final notice on parallelism/distribution issues of the algorithm, it is true that in order for the algorithm to start processing any itemset of size $k$ all itemsets of smaller size must have been processed first; therefore, one could argue, during the processing of the last itemsets of a particular size, it is possible that only a small number of CPU cores will be utilized due to lack of work to do (larger itemsets cannot start their processing before all current itemsets are processed). However, given a large enough dataset (in terms of number of user histories) it is still possible to maintain essentially 100% utilization of all processing units due to the fact that the computation of a rule's support and interestingness values is where QARMA spends the vast majority of its execution time, and are inherently fully parallelizable operations for all interestingness metrics we care about (i.e. confidence, conviction, lift, leverage.)

*2.2.1 Example Run of QARMA*

To illustrate the workings of the algorithm, we follow its trace of execution on a trivially small dataset. Consider a dataset $D_{ex}$ containing the histories of purchase transactions of 6 users with user-ids 1,2,…6 having purchased 3 items (with item-ids labeled as $a, b, c$) each:

- the 1st user with userid=1 has the following history $\{(IID = a, p = 1.0, p^- = -1.0), (IID = b, p = 1.0, p^- = -1.0), (IID = c, p = 0.3, p^- = -0.3)\}$.
- The 2nd user (userid=2) has the following history $\{(IID = a, p = 1.0, p^- = -1.0), (IID = b, p = 1.0, p^- = -1.0), (IID = c, p = 0.2, p^- = -0.2)\}$.
- The 3rd user with userid=3 has the history $\{(IID = a, p = 1.0, p^- = -1.0), (IID = b, p = 1.0, p^- = -1.0), (IID = c, p = 0.1, p^- = -0.1)\}$.
- The 4th user has history $\{(IID = a, p = 1.0, p^- = -1.0), (IID = b, p = 0.6, p^- = -0.6), (IID = c, p = 0.3, p^- = -0.3)\}$.
- The user with userid=5 has history $\{(IID = a, p = 0.9, p^- = -0.9), (IID = b, p = 0.5, p^- = -0.5), (IID = c, p = 0.2, p^- = -0.2)\}$.
- And the final 6th user has history $\{(IID = a, p = 0.8, p^- = -0.8), (IID = b, p = 0.5, p^- = -0.5), (IID = c, p = 0.2, p^- = -0.2)\}$.

In this dataset, besides the common shared price attribute $p$, in order to be able to derive rules specifying closed intervals for the price attributed in the antecedents, we also include the (artificial) attribute $p^-$ that represents the negative value of the price at which the item was purchased; see discussion in section 1. We only show the execution of the algorithm steps 3.2.2–3.2.3 and the quantitative rules produced with the qualitative rule "$a \to b$" (with support for the itemset $(a, b)$ being 100% as all users have in their histories both items.) We assume that the minimum required support $s = 0.4$ and that the sole metric of interestingness is confidence with a minimum required threshold $c = 0.7$. We also assume a lexicographic ordering function for the items, and that the price attribute for each item is ranked first so that the negated-price attribute is ranked second.

In step 3.2.2.2.1, the tuple $Q = (b, p, 1.0)$ corresponding to the rule $a \to b[p \geq 1.0]$ has a support of 0.5, so it is added in step 3.2.2.2.3 onto $T$. The elements that the for-loop of step 3.2.2.2.4.2 iterates over, is the single antecedent $\{a\}$ which is not yet quantified in $Q$. The check in step 3.2.2.2.4.2.0 fails, so execution continues with the iterations over the attributes of item $a$ in the for-loop of step 3.2.2.2.4.2.1, with the price attribute being checked first, and the negated-price attribute checked second.

In the first iteration, when the price attribute is checked, it is clear that neither the condition $ord(a) < max_{A \in \{a\}}\{ord(A): \exists v: (A, p, v) \in Q\}$ holds nor the condition $ord(p, a) \leq max_{t \in \{p, p^-\}}\{ord(t, a): \exists v: (a, t, v) \in Q\}$ holds in step 3.2.2.2.4.2.1.1, so the execution proceeds with the loop over all possible values that the price attribute takes for the item $a$, which are the values of the components of the vector $P(a, p) = [0.8 \; 0.9 \; 1.0]^T$.

In step 3.2.2.2.4.2.1.2.1 the set $Q' = \{(a, p, 0.8), (b, p, 1.0)\}$ corresponding to the quantitative rule $a[p \geq 0.8] \to b[p \geq 1.0]$ is formed, and since its support is 0.5, it enters $T$ in step 3.2.2.2.4.2.1.2.2.1. However, because the confidence of this rule is $0.5 < c$ the check for interestingness in steps 3.2.2.2.4.2.1.2.2.2 fails, and execution continues with the next price level for item $a$ which is the value 0.9. Now, the set $Q' = \{(a, p, 0.9), (b, p, 1.0)\}$ is formed, corresponding to the rule $a[p \geq 0.9] \to b[p \geq 1.0]$. This new rule has support 0.5 (thus enters $T$), and confidence 0.6 which is still short of the required threshold confidence level. Now, the last price level for item $a$ is tried which gives the set $Q' = \{(a, p, 1.0), (b, p, 1.0)\}$ corresponding to the rule $a[p \geq 1.0] \to b[p \geq 1.0]$. This rule has support 0.5 (thus enters $T$), and confidence 0.75, which is above the minimum required threshold, thus in step 3.2.2.2.4.2.1.2.2.1.1 this rule becomes the first rule to enter the set $R_{local}$.

Then, execution of the for-loop over the attributes of item $a$ (step 3.2.2.2.4.2.1) continues with the negated-price attribute, whose levels are the values of the components of the vector $P(a, p^-) = [-1.0 \; -0.9 \; -0.8]^T$. In step 3.2.2.2.4.2.1.1, the set $Q = (b, p, 1.0)$ and the tests in step 3.2.2.2.4.2.1.1 again fail for item $a$ and attribute $p^-$ so execution proceeds with the loop over the values of the vector $P(a, p^-)$ for this item. The first quantification $Q' = \{(a, p^-, -1.0), (b, p, 1.0)\}$ corresponds to the rule $a[p \leq 1.0] \to b[p \geq 1.0]$ whose support is 0.5 (thus enters $T$), but its confidence is 0.5, so does not enter $R_{local}$. The second quantification $Q' = \{(a, p^-, -0.9), (b, p, 1.0)\}$ has 0.0 support and so further processing of the values of the negated-price attribute of item $a$ is stopped at this point due to the break in step 3.2.2.2.4.2.1.2.3. Since there are no other attributes to loop over in step 3.2.2.2.4.2.1 and also the rule has no other antecedents to check in step 3.2.2.2.4.2, control is transferred to step 3.2.2.2.4.

Now, the second partial quantification that entered the set $T$, namely the set $Q' = \{(a, p, 0.8), (b, p, 1.0)\}$ is retrieved as $Q$ in step 3.2.2.2.4.1 and the test in step 3.2.2.2.4.2.0. $(ord(a) < max_{L \in \{a\}, t \in A_Q^L}\{ord(L): \exists v: (L, t, v) \in Q\})$ fails, so control continues with the execution of the for-loop over the attributes of item $a$ in 3.2.2.2.4.2.1; the test for the first (price) attribute in step 3.2.2.2.4.2.1.1 succeeds as $ord(p, a) = max_{t \in \{p, p^-\}}\{ord(t, a): \exists v: (a, t, v) \in Q\}$ and the algorithm proceeds to examine the values for the negated-price attribute $p^-$.

Next the rule $a[p \geq 0.8][p \leq 1.0] \rightarrow b[p \geq 1.0]$ is examined, and with support 0.5, the corresponding $Q'$ enters $T$ but as its confidence is 0.5, is not added to $R_{local}$. The rule $a[p \geq 0.8][p \leq 0.9] \rightarrow b[p \geq 1.0]$ is examined but as its support is 0, processing of this attribute breaks again. The next $Q$ to be retrieved from the set $T$ is the set $Q' = \{(a, p, 0.9), (b, p, 1.0)\}$, and, as before, the values of the negated-price attribute of item $a$ are examined. First, the rule $a[p \geq 0.9][p \leq 1.0] \rightarrow b[p \geq 1.0]$ is examined, which has support 0.5 (thus the corresponding $Q'$ enters $T$) but as the confidence is only 0.6, is not added to $R_{local}$. The rule $a[p \geq 0.9][p \leq 0.9] \rightarrow b[p \geq 1.0]$ (which is equivalent to the rule $a[p = 0.9] \rightarrow b[p \geq 1.0]$) is examined then, but because support is 0, processing again breaks.

Next, the rule $a[p \geq 1.0][p \leq 1.0] \rightarrow b[p \geq 1.0]$ is examined, and the rule has support 0.5 (thus is added onto $T$) but even though its confidence is 0.75, is *not* added onto $R_{local}$ because the rule is dominated by the rule $a[p \geq 1.0] \rightarrow b[p \geq 1.0]$ which is already in $R_{local}$.

At this point, all the other $Q$ elements currently in the set $T$ fail to pass the tests within the for-loop of step 3.2.2.2.4.2 and the first major iteration of the algorithm over the price levels of the consequent item $b$ has concluded (now $T = \emptyset$), and the next iteration begins, with price level for the consequent item 0.6. The first rule tried is $a \rightarrow b[p \geq 0.6]$ which has support 0.6 and enters $T$ but its confidence is 0.6, less than minimum required threshold to enter $R_{local}$. The same with the quantification $a[p \geq 0.8] \rightarrow b[p \geq 0.6]$. But the next quantification $a[p \geq 0.9] \rightarrow b[p \geq 0.6]$ does meet both support and confidence targets, and enters $R_{local}$ as well (is not dominated by previous rule because of lower price level for the antecedent, as well as higher support and confidence levels). Similarly, the next tried rule $a[p \geq 1.0] \rightarrow b[p \geq 0.6]$ meets required criteria, and enters $R_{local}$ as it is not dominated by the previous rule that entered the set $R_{local}$ due to higher confidence. Quantifying the negated-price attribute again provides no valid *and* non-dominated rules, so the algorithm enters its final major iteration over the price levels of the consequent item, with price value for $b$ the number 0.5. In this case, the only valid, non-dominated rule found is $a[p \geq 0.8] \rightarrow b[p \geq 0.5]$ with support and confidence 1.0. Thus the total valid non-dominated rules found from the base rule $a \rightarrow b$ are:

1. $a[p \geq 1.0] \rightarrow b[p \geq 1.0]$ with support=0.5, confidence=0.75
2. $a[p \geq 0.9] \rightarrow b[p \geq 0.6]$ with support=0.667, confidence=0.8
3. $a[p \geq 1.0] \rightarrow b[p \geq 0.6]$ with support=0.667, confidence=1.0
4. $a[p \geq 0.8] \rightarrow b[p \geq 0.5]$ with support=1.0, confidence=1.0

The coverage of the dataset by the above rules is 100%. For completeness purposes, we mention that there are in total 17 valid non-dominated rules that can be derived from this dataset, and they all have only one item in their antecedents list.

## 3 COMPUTATIONAL RESULTS

Throughout this section, we use $CONFIDENCE(r)$ as the sole interestingness metric for our algorithm, and consequently, we use $LTF(r, r') = CONFIDENCE(r) \leq CONFIDENCE(r')$ as our predicate formula in the definition of the dominance relation between two rules. We selected confidence as our interestingness metric because this is the most widely used metric in the literature.

### 3.1 Application Domains
We test the performance of our algorithm in three domains.
- The first domain is that of *e–commerce*, where in the absence of real–world data, we have designed and implemented a synthetic dataset generator that simulates a large–scale marketplace.
- The second domain of application is that of *rare event detection in predictive maintenance settings*. Again, in the absence of publicly available real-world data, we have designed and implemented a synthetic dataset generator that simulates rare events in a multi-dimensional setting that indicate the need for maintenance.
- The third domain of application is that of *recommender systems*; for this domain, we have tested the accuracy of the rules produced by QARMA in predicting the movie ratings in the 1mio version of the Movie–Lens dataset (ml-1m) [26], and we have also separately designed and implemented a post–processor algorithm that re–ranks the results of the recommendation algorithms we have used in the commercial system AMORE [12] on a real–world dataset. All results indicate that QARMA has superior performance that translates directly to improved recall values for recommender systems, and into significant revenues increase for the simulated e–commerce scenario. We compare QARMA against two well-known systems for mining quantitative association rules, namely QuantMiner [17] and MGR [27], and we show that our algorithm compares very favorably against those systems. We also test our system against the quantitative mining versions of the algorithms A–Priori and Eclat as found in the Open–Source Software KEEL (www.keel.es) [28]. KEEL discretizes the numeric attributes of a dataset in up to 9 different equi-width partitions, and works with the resulting categorical–attributes–only dataset. Our results, again, compare very favorably against those systems as well.

### 3.2 E-commerce Scenario Synthetic Dataset Generation & Results

As QARMA can find applicability in many different e–commerce related scenarios, we performed a series of simulations that allowed the evaluation of the performance of the algorithm under different conditions. Toward this end, we designed and implemented a Synthetic Dataset Generator (SDG), a program that allows the generation of datasets according to specific predefined configuration parameters that simulate the operations of a real–world market.

For our base-line, we assume a constant, user–defined number of (rational) consumers, and a store (marketplace) having sufficient stock from a constant list of user–specified items, some of which are specified as being of elastic demand, and others having inelastic demand. Our simulations run for a number of user–defined cycles, representing consumers' "weekend shopping trips". At the beginning of each cycle, each simulated consumer makes up a wish–list of items to look for, according to a Pareto–law distribution, which models the well–known phenomenon that in each category of consumer goods, a few items become very popular whereas the majority of items lag significantly behind the best–sellers. Repeat purchases of the same item are allowed, to model consumable products.

Inelastic demand items that end up in the wish–list of a consumer are always bought by the consumer regardless of their current price tag. On the contrary, elastic demand items on the wish-list of a consumer will only be purchased if the current price tag of the item is below the consumer's "*reservation price*" for that item, a value a–priori specified (randomly) separately for each consumer and each elastic–demand item.

After running the simulation for a number of pre–specified cycles, the transaction histories of all simulated consumers are recorded in a database, and we have generated our synthetic commerce dataset. For our experiments, we have chosen three different major configurations; we report results from the 1st configuration where there are 2000 items, 51% of which are elastic demand items, 2000 users, and we run the simulation for 10 cycles, where each user makes 10 purchases per cycle.

After running QARMA, we ran a validation test to see whether the QARMA produced rules can be used to increase retailer revenues by targeted discounts. To this end, we ran QARMA on the dataset produced by the 1st configuration mentioned above, and then ran again the synthetic dataset generation algorithm three more times, for another 100 cycles. The first run of the synthetic dataset generation was executed without any modifications, starting exactly from the last cycle where the original execution had finished, the second run was executed with the added modification that all items were on a fixed percent sale (5, 10, 15, … 40% discounts), and the third and last run was executed with the modification that the system was allowed to lower the price of each item to the individual customer's estimated reservation price as long as this price discount was less than the same fixed percent sale (5, 10, … 40%) that now represented simply the maximum sale discount that the retailer would be willing to offer to any particular customer.

For our runs, we estimated a customer's $c$ reservation price $p_{c,i}$ for an item $i$ to be the *largest* quantitative value of the consequent among all rules whose confidence in the training set was above 70%, produced from QARMA that "*fired*" for that customer having that specific item in their consequent attribute. The results were tested for statistical significance and pass the *Student t–test*, and also the *sign* as well as the *signed rank test* at the 95% confidence level. They are shown in Table I and they verify that QARMA produces rules that can be used to significantly enhance retailer revenues, as the sales become personalized rather than horizontal discounts across all customers. As expected, as QARMA is allowed to give higher discounts to targeted customers, the revenues increase over the base line. The opposite however is true when discounts are applied horizontally, i.e. when discounts are offered indiscriminately, the higher the discount level, the higher the loss of revenue for the retailer.

10TABLE I: RETAILER REVENUE INCREASE BY INCORPORATION OF QARMA RESULTS.

| SDG#1 Conf.Thres.=0.7, Supp.Thres.=0.35 | Sales | | | | |
|---|---|---|---|---|---|
| Discount Level % | Base Level (No Discounts) | Fixed Horizontal Discounting | Fixed Discount Revenue Increase % | QARMA-based Discounting | QARMA-based Discount Revenue Increase % |
| 5% | 853160 | 810502,00 | -5,00% | 896838,00 | 5,12% |
| 10% | | 780754,50 | -8,49% | 896838,00 | 5,12% |
| 15% | | 757184,25 | -11,25% | 896838,00 | 5,12% |
| 20% | | 728248,80 | -14,64% | 896838,00 | 5,12% |
| 25% | | 698847,00 | -18,09% | 896838,00 | 5,12% |
| 30% | | 661693,20 | -22,44% | 896838,00 | 5,12% |
| 35% | | 626696,85 | -26,54% | 897410,00 | 5,19% |
| 40% | | 592685,39 | -30,53% | 897851,00 | 5,24% |

### 3.3 Rare-Event Detection in Predictive Maintenance Scenario: Synthetic Dataset Generation & Results

Predictive maintenance is now recognized as an important enabler technology for smart industry, also known as "industry 4.0" with the potential to significantly improve various key performance indicators. Predictive maintenance replaces or at least complements traditional scheduled maintenance suggested by the manufacturer of a machine or tool, by monitoring the machine and its environment via a large number of sensors and then learning to detect the events that trigger the need for maintenance of the machine. Such events are of course rarely occurring in the life-time of the machine, and therefore there is a strong need for algorithms that can learn to detect rare events with high accuracy and very low false alarm rates.

The QARMA framework is theoretically at least ideally suited for such a task because it can automatically discover all non-dominated rules that indicate that a set of (rarely occurring) pre-conditions triggers a maintenance need. By specifying sufficiently low support levels so as to capture the rarity of the events occurring frequency, and at the same time specifying sufficiently high required confidence and lift or conviction values for the derived rules, the algorithm can produce a (possibly large) set of rules that trigger rarely, but when they do, they indicate with high likelihood the need for maintenance. Once produced, such rules are continually and automatically tested to see if they fire as new observations arrive. When they fire, they signal an alert for a maintenance need.

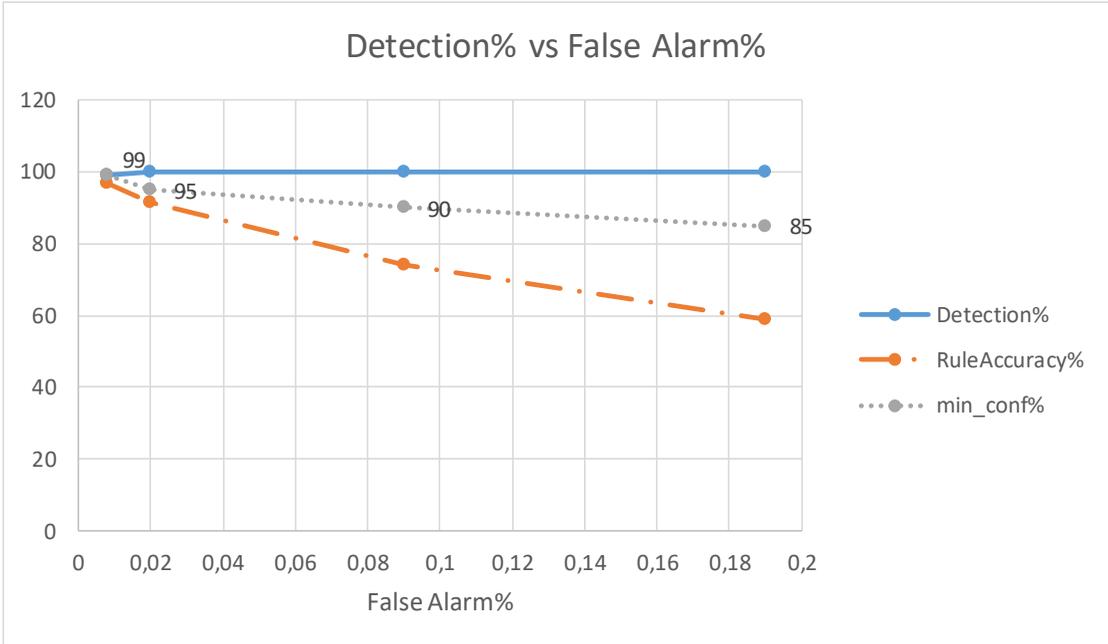

Fig. 1: Detection of rare events in synthetic datasets using QARMA with small required support (0.15%). When the minimum required confidence is set at 99%, the rules produced by running QARMA on a training set detect 99% of all rare events in the test-set, with a 0.02% false-alarm rate, and an overall rule accuracy on the test-set of 97.05%. The Area-Under-the-Curve is practically 1. The running time of QARMA on an intel i7 CPU was less than one minute.

To test the ability of QARMA to detect such rare events in practice, we have designed a testbed generator that creates random multi-dimensional data-points (in $\mathbb{R}^{20}$) from many Gaussian distributions. Each dimension value of each point may be missing with a probability of 90%, and thus the generated dataset is 90% sparse. On top, we generate a very small number of points whose attribute values for a small number of dimensions (by default the last 3) lies within 1% of their extremal values; each of these points is "tagged" as indicating the need for maintenance, by drawing a "class" attribute value from the distribution $N(50, 10^2)$ as opposed to the distribution $N(0,1)$ that is chosen for the "normal" instances. We generate 35000 training data points in total (with an extra 100 data points indicating the need for maintenance); an equally sized and structured dataset is then generated for testing purposes. We run QARMA on the training set asking for minimum rule support of 0.15% and confidence ranging between 85% and 99% on the training dataset, and then apply the produced rules onto the reserved test set to see the detection rate versus the false alarm rate of the produced rules. The results shown in fig. 1 completely justify the use of QARMA as a promising tool for detecting rare events in predictive maintenance settings.

### 3.4 Movie-Lens Results

We also test the accuracy of QARMA results on the well–known Movie–Lens dataset ml-1m, where we treat the movies in the dataset as the items, and the rating of a movie given by a user (in an integer scale from 1 to 5) to be the quantitative attribute of each item. Our goal is to derive quantitative rules that can predict the minimum rating a user would give to a movie from their transactional history (i.e. the ratings they have provided for other movies). We only use a single quantitative attribute for all items, namely the rating provided by the users. We compare the performance of QARMA against the Movie–Lens ml-1m dataset. The ml-1m dataset contains 1.000.209 ratings from 6040 different users on 3706 different movies.

Table III shows the run-times of QARMA when using between 1 and 8 threads in an intel core i7 CPU on a Lenovo ThinkPad T530 with 16GB RAM running Windows 10. This particular CPU is a quad-core processor with hyper-threading hardware technology, so that the O/S sees 8 logical cores. The first column shows the number of machines used, the second the threads QARMA spawns, the third column shows the response time of the entire QARMA algorithm, and the fourth column shows the number of seconds elapsed in the computation of the frequent itemsets of length up to 3 (which is a sequential operation). In the last two rows of this table, we show results when running the same experiment, on a cluster of 2 or 3 machines equipped with an Intel core i7 CPU and 8GB RAM each (for a total number of 16 and 24 threads respectively), connected via a 100Mbps LAN. For this experiment on a cluster parallel distributed architecture, we utilize the distributed batch task execution software architecture implemented in the "`parallel.distributed`" package of the popt4jlib java library, which offers the advantage of elastic computing in that machines may be added or removed from the cluster while QARMA is running (hot-plugging).

In fig. 2a we show the speed-up and efficiency of the *parallel* computations of QARMA; the graphs show that QARMA scales linearly (or even super-linearly) in the number of threads, *up to the number of available real cores of the i7 CPU*. The super-linear scaling can be attributed to less context-switching, cache misses or cache movements as the number of available cores increases. The observed super-linearity is maintained for the parallel computing part of the QARMA algorithm when the distributed computing environment mentioned above is utilized. In fig. 2b, we show the speed-up and efficiency of the *distributed* version of QARMA. In this case, we attribute the super-linear speed-up of the algorithm in the existence of hyper-threading in each of the CPUs utilized: in particular, as we spawn 8 threads on each quad-core machine with hyper-threaded CPUs, we indeed observe higher than linear speed-up of the execution time of the parallel computations of QARMA.

An important implementation detail regarding the parallelism of QARMA is the following: for the ml-1m dataset, the running time required to run steps 3.2.2.1—3.2.2.3 of any candidate rule is very short —because of the very small number of different ratings possible. Therefore, when submitting each frequent itemset as a separate task to be processed in parallel with all others of the same length, the overhead incurred by the locking required by the associated thread–pool queue become significant. In this case, "*batching*" helps: for example, when aggregating 128 rule processing tasks (representing rules of same length) together as a single task to be submitted to the thread–pool, the running time of QARMA when it spawns 8 threads on a single i7 CPU, drops from 595.5 seconds shown in Table II, to 345.7 seconds!

TABLE II: QARMA RESPONSE TIMES ON ml-1m DATASET FOR VARIOUS NUMBER OF THREADS AND MACHINES.

| #Machines | #Threads | total response-time (secs) | itemset creation time (secs) |
|---|---|---|---|
| 1 | 1 | 2382,7 | 196 |
| 1 | 2 | 1085,9 | 206 |
| 1 | 3 | 837,5 | 198 |
| 1 | 4 | 736 | 196 |
| 1 | 5 | 674,7 | 183 |
| 1 | 6 | 670,1 | 181 |
| 1 | 7 | 592,2 | 185 |
| 1 | 8 | 595,5 | 202 |
| 2 | 16 | 430 | 210 |
| 3 | 24 | 353 | 208 |

Table III compares the performance of QARMA (using all 8 logical cores of the CPU) against the Apriori-A and Eclat-A algorithm implementations found in the KEEL machine learning library, the MGR algorithm (implemented in Java by the second author of this paper, since no Open Source version is available), and the Genetic Algorithm implementation of the QuantMiner Open Source system. For the ml-1m dataset, we assume two quantitative attributes, namely the ratings of each item, plus the negation of the rating (so as to produce rules whose antecedents allow item ratings to be restricted in fully closed intervals). The third and fourth column show the minimum support and confidence thresholds in the experiments, the fifth column describes the number of rules found by each algorithm, the sixth the response time of the algorithm in seconds, the seventh column the dataset coverage of the rules found, and the final column describes the number of rules found by the other algorithms that are dominated by one or more of the rules that QARMA produced. The Eclat-A algorithm ran out of memory after the time shown in the specified column; the GA in the QuantMiner system was not able to start due to a limitation in the total number of attributes to consider (the ml-1m Movie–Lens dataset has a total of 3706 attributes). For these runs, we restricted QARMA processing to all frequent itemsets of total length up to three; frequent itemsets were computed using the FPGrowth algorithm implementation found in the University of Waikato's WEKA library. The results show the clear superiority of QARMA in terms of the solution found, both in terms of the number of rules found, and of dataset coverage; in fact, none of the other systems was able to find any rule that wasn't dominated by a rule that QARMA found. And even in terms of response times, QARMA compares very well with all the other systems in the comparison suite, except the numeric version of the A–Priori algorithm, which however manages to find an almost trivial set of rules from the dataset, with a significant number of the rules found being redundant, in that they were dominated by other rules in the result set.

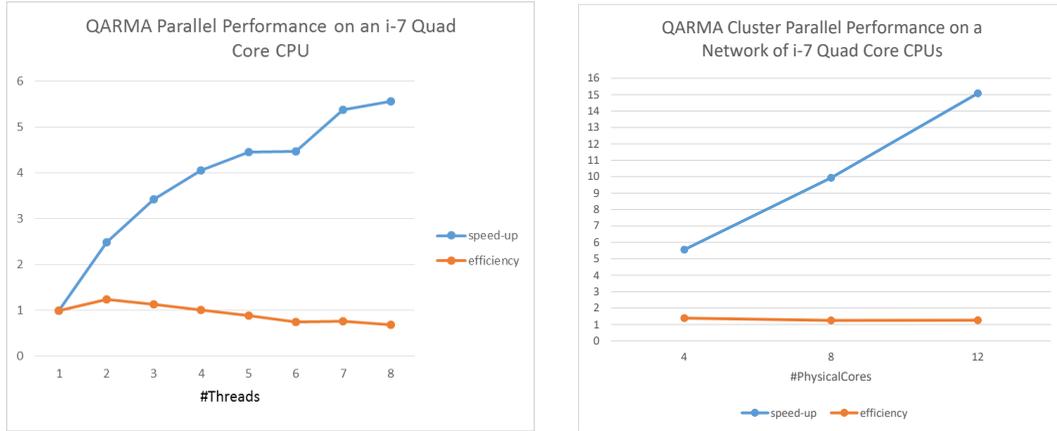

Fig. 2: (a) Speed-up and efficiency of QARMA tested on the ml-1m dataset. (b) Speed-up and efficiency of QARMA tested on the ml-1m dataset on a network of up to 3 i7 quad core CPUs; the #cores shown refers to physical ones.

TABLE III: EXPERIMENTS WITH MOVIE–LENS ML-1M DATASET

| Algorithm | Dataset | s | c | #rules | run-time (secs) | coverage% | #dominated rules by QARMA |
|---|---|---|---|---|---|---|---|
| QARMA | ML_1M | 0.05 | 0.8 | 459161 | 19512.7 | 99.0% | - |
| Eclat-numeric | | | | - | 69771.1 | - | N/A |
| A-priori-numeric | | | | 52 | 39.8 | 18.5% | 52(self: 16) |
| MGR | | | | 1 | 6261.0 | 21.6% | 1 |
| QuantMiner | | | | - | - | - | N/A |
| QARMA | | 0.1 | 0.8 | 40513 | 345.7 | 96.3% | - |
| Eclat-numeric | | | | - | 69677.0 | - | N/A |
| A-priori-numeric | | | | 4 | 31.8 | 38.4% | 4 |
| MGR | | | | 0 | 302 | 0.0% | N/A |
| QuantMiner | | | | - | - | - | N/A |

A few sample rules produced by QARMA on ml-1m when $s_{min} = 0.05$ and $c_{min} = 0.8$ are the following:
- "Heat (1995)"≥1 AND "Twelve Monkeys (1995)"≥4 → "Usual Suspects (1995)"≥4 [supp=5%, conf=80.5%, conv=376.9%, lift=3.0]
- "Usual Suspects (1995)"≥1 AND "Taxi Driver (1976)"≥1 → "Fargo (1996)"≥3 [supp=10.4%, conf=90.1%, conv=617.1%, lift=2.3]
- "Batman forever (1995)"≥2 AND "Star Wars Episode VI – Return of the Jedi (1983)"≥1 → "Raiders of the Lost Arc (1981)"≥4 [supp=6.9%, conf=80.8%, conv=326.1%, lift=2.2]

In the list above, "supp" stands for the support of the rule (Defn. 2), "conf" stands for rule confidence (Defn. 1), "conv" stands for rule conviction (Defn. 3), and "lift" is clearly the lift of the rule, defined as the confidence of the rule divided by the number of user histories in which the consequent occurs in a transaction with a value for the common attribute $p$ greater than or equal to the value specified by the rule's consequent.

## 4 CONCLUSIONS & FUTURE DIRECTIONS

We presented QARMA, a novel highly parallel distributed algorithm for deriving all quantitative association rules in datasets containing many quantitative attributes, constrained to intervals describing half–lines, and with the help of special attributes, closed intervals. We have shown that QARMA is particularly well-suited for many diverse applications. We conducted computational experiments on large datasets demonstrating the scalability of the algorithm, and have compared its performance against several other algorithms for quantitative association rule mining.

In the near future, we plan to extend the QARMA algorithm so as to further develop heuristics that significantly speed up the algorithm without any serious sacrifices on the completeness or soundness properties of the algorithm; and also, to more tightly integrate the qualitative association rule mining phase (step 1 of the algorithm) with the rest of the algorithm, which could also speed up the algorithm even further.

Incorporating online learning (also known as continuous learning) techniques into QARMA so as to allow incremental processing of the dataset as more user histories are added into the data, or as more data are added to existing user histories, without the need to process from scratch the entire dataset, is another very important and exciting future line of work.

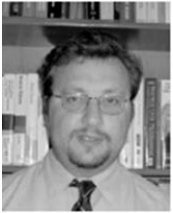

**Ioannis T. Christou** holds a Dipl. Ing. Degree in Electrical Engineering from the National Technical University of Athens (NTUA), Greece (1991), an M.Sc. (1993) and Ph.D. (1996), both in Computer Sciences from the University of Wisconsin at Madison (Madison, WI, USA), and an MBA joint degree from NTUA and Athens University of Economics and Business (AUEB) (2006). He has been with Delta Technology Inc. as a Senior Developer, with Intracom S.A. Development Programmes Dept. as an Area Leader in Data & Knowledge Engineering, and with Lucent Technologies Bell Labs as a Member of Technical Staff. He has also been adjunct professor at Carnegie-Mellon University, Pittsburgh, PA, USA. He is currently Professor at Athens Information Technology, Athens, Greece, heading the Big Data Mining Group, and has published more than 70 articles in journals and peer-reviewed conferences. His current research interests include Analytics & Computational Intelligence, Optimization, Data Mining, and Parallel & Distributed Computing. Dr. Christou is a member of the Technical Chamber of Greece.

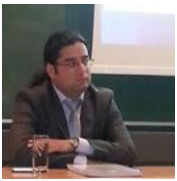

**Emmanouil Amolochitis** received the B.Sc. degree in Information Technology in 2005 from Deree, The American College of Greece, the M.Sc. degree in Information Technology & Telecommunications in 2009 from Athens Information Technology, Greece, and the Ph.D. degree from CTiF, at Aalborg University, Denmark in 2014. He has held positions as software engineer at Intrasoft S.A. and Ethno-data, and as research engineer at Voice Web S.A. He is currently data scientist at Cepal Financial Services S.A. Greece. His interests include Data Mining, Machine Learning and Information Retrieval.

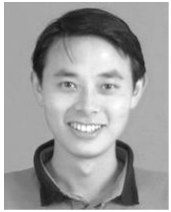

**Zheng-Hua Tan** received the B.Sc. and M.Sc. degrees, in 1990 and 1996, respectively, both in Electrical Engineering from Hunan University, China, and the Ph.D. degree in Electronic Engineering from Shanghai Jiao Tong University, China, in 1999. He is professor in the Dept. of Electronic Systems at Aalborg University, Denmark. Before joining AAU, he was a postdoctoral fellow in the Dept. of Computer Science at the Korea Advanced Institute of Science and Technology (KAIST), Korea, and was also a visiting scientist at the Computer Science and Artificial Intelligence Laboratory, Massachusetts Institute of Technology, USA. He was also an associate professor in the Dept. of Electronic Engineering at Shanghai Jiao Tong University. His interests include Computational Intelligence, Speech Recognition and Machine Learning. He has published extensively in these areas in refereed journals and conference proceedings, and is a member of the editorial board of many journals. He is a senior member of the IEEE.